\documentclass[letterpaper]{article} 
\usepackage{aaai24}  
\usepackage{times}  
\usepackage{helvet}  
\usepackage{courier}  
\usepackage[hyphens]{url}  
\usepackage{graphicx} 
\usepackage{natbib}  
\usepackage{caption} 
\usepackage{algorithm}
\usepackage{algorithmic}
\usepackage{amsmath}
\usepackage{subfigure}
\usepackage{amssymb}
\usepackage{pifont}
\usepackage{multirow}
\DeclareMathAlphabet\mathbfcal{OMS}{cmsy}{b}{n}

\usepackage{newfloat}
\usepackage{listings}
\DeclareCaptionStyle{ruled}{labelfont=normalfont,labelsep=colon,strut=off} 
\floatstyle{ruled}
\newfloat{listing}{tb}{lst}{}
\floatname{listing}{Listing}
\title{Temporal Adaptive RGBT Tracking with Modality Prompt}
\author {
    Hongyu Wang,
    Xiaotao Liu\thanks{Xiaotao Liu is the corresponding author.},
    Yifan Li,
    Meng Sun,
    Dian Yuan,
    Jing Liu
}
\affiliations {
    Guangzhou Institute of Technology, Xidian University, Guangzhou, China\\
    22171214782@stu.xidian.edu.cn, xtliu@xidian.edu.cn, 
    18066899461@163.com, sun\_meng2002@163.com,
    d1anskr@163.com, neouma@163.com
}

\usepackage{bibentry}

\begin{document}

\maketitle

\begin{abstract}
RGBT tracking has been widely used in various fields such as robotics, surveillance processing, and autonomous driving. Existing RGBT trackers fully explore the spatial information between the template and the search region and locate the target based on the appearance matching results. However, these RGBT trackers have very limited exploitation of temporal information, either ignoring temporal information or exploiting it through online sampling and training. The former struggles to cope with the object state changes, while the latter neglects the correlation between spatial and temporal information. To alleviate these limitations, we propose a novel Temporal Adaptive RGBT Tracking framework, named as TATrack. TATrack has a spatio-temporal two-stream structure and captures temporal information by an online updated template, where the two-stream structure refers to the multi-modal feature extraction and cross-modal interaction for the initial template and the online update template respectively. TATrack contributes to comprehensively exploit spatio-temporal information and multi-modal information for target localization. In addition, we design a spatio-temporal interaction (STI) mechanism that bridges two branches and enables cross-modal interaction to span longer time scales. Extensive experiments on three popular RGBT tracking benchmarks show that our method achieves state-of-the-art performance, while running at real-time speed.
\end{abstract}

\section{Introduction}

As a fundamental task in computer vision, visual object tracking (VOT) aims at localizing a specified object in each video frame given its initial state. Thanks to the efforts of researchers, many excellent works \cite{bertinetto2016fully,bhat2019learning,chen2021transformer,cui2022mixformer} have been proposed. Despite the promising results, limited by imaging mechanisms of visible images, RGB-based trackers struggle to achieve good performance in some complex scenarios, such as extreme illumination and bad weather. The thermal infrared images (TIR or T) are insensitive to these factors, which can provide complementary information to visible images \cite{zhang2020object}. As a result, RGBT tracking has become a popular research topic in recent years. It has been successfully deployed in various applications such as robotics \cite{chen2017rgb}, surveillance processing \cite{alldieck2016context}, and autonomous driving \cite{dai2021tirnet}.

In RGBT tracking, both spatial and temporal information are crucial for target localization. The former contains object appearance information and the latter contains the state changes of objects among frames. The first consideration for RGBT tracking is the fusion of complementary information from RGB and TIR modalities, which is mainly a spatial interaction. Temporal information is easy to be neglected when designing high-performance modality fusion tracking algorithms. However, without proper exploitation of temporal information, the tracker can hardly cope with some complex scenarios (e.g., aspect ratio changes, target deformation, and fast movement,).
 \begin{figure}[!t]
    \centering
    \includegraphics[width=1.0\linewidth]{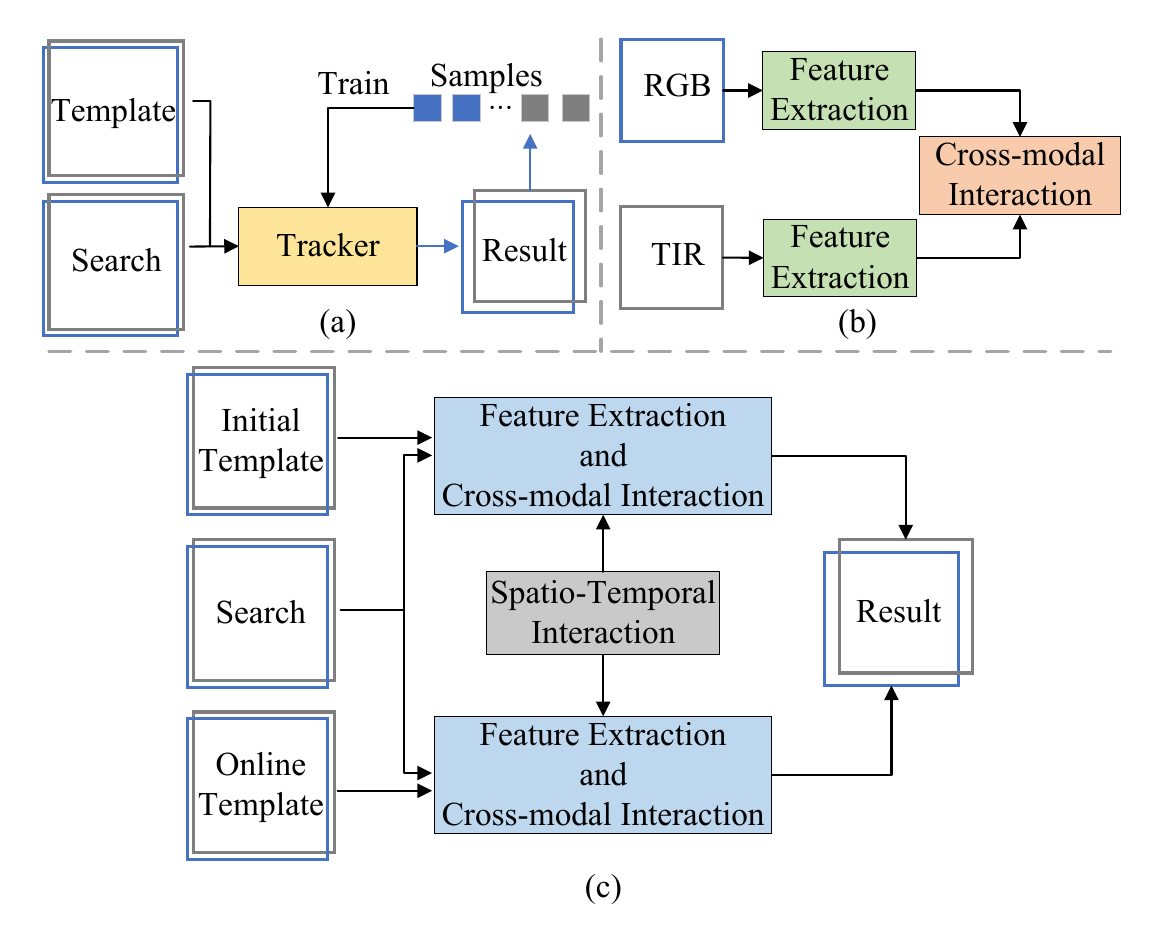}
    \caption{Differences between our RGBT tracking approach and previous ones. (a) Sampling from tracking result and training online. (b) Processing RGB and TIR images separately and performing cross-modal interaction through specialized networks. (c) Performing feature extraction and cross-modal interaction simultaneously and capturing temporal information by online updated templates.}
    \label{fig:1}
    \end{figure}

According to the exploitation way of temporal information, the prevailing RGBT trackers can be divided into the following two categories. The first one neglects temporal information, which only refers to the initial template to build an appearance model and find the best match in subsequent frames \cite{zhang2019siamft,zhu2021rgbt,peng2023siamese,feng2022learning}. Offline appearance model without considering the state changes of objects leads to limited discriminative ability. The second one uses the tracking results of each frame to generate new training samples and trains the tracker online during tracking process \cite{wang2018learning,zhai2019fast,zhang2018learning,mei2023differential}, as shown in Figure \ref{fig:1} (a). However, this update method is prone to tracking drift and neglects the correlation of spatial and temporal information. In addition, the tracking results are not always reliable. The updates when tracking fails can even make the tracker lose discriminative ability. Thus, can we find a more effective manner to exploit temporal information?

While exploiting temporal information, we need to ensure full cross-modal interaction for each pair of images (from RGB and TIR modalities). Most RGBT trackers usually use different branches to process images in RGB and TIR modalities separately, then perform cross-modal interaction through specialized networks, as shown in Figure \ref{fig:1} (b). This paradigm is prone to insufficient cross-modal interactions. Recently, in Natural Language Processing (NLP) field, prompt-tuning \cite{lester2021power} has become the dominant paradigm, which adapts the foundation model to different tasks by adding a textual prompt to the model inputs. Some researchers \cite{bahng2022exploring,jia2022visual,radford2021learning,zheng2022prompt} have transferred this paradigm to computer vision by adding learnable visual prompts to the frozen base model. Given the large inheritance between VOT and RGBT, some works \cite{yang2022prompting,zhu2023visual} add modality prompts to the frozen RGB tracker for RGBT tracking and achieve comparable performance to full fine-tuning. We observe that modality prompt achieves the information complementation of RGB and TIR modalities in a simple and effective way. Inspired by this, we integrate feature extraction and cross-modal interaction with the help of modality prompts.

Based on the above analysis, we propose a novel temporal adaptive RGBT tracking framework with a spatio-temporal two-stream structure, named as TATrack. Different from common two-branch networks that process images in RGB and TIR modalities separately, we simultaneously perform feature extraction and cross-modal interaction in each branch and capture temporal information by online updated templates, as shown in Figure \ref{fig:1} (c). Compared with previous methods, this processing paradigm has several advantages as follows. First, the tracker can fully learn the modality fusion by integrating multi-modal feature extraction and cross-modal interaction. Second, it enables model to select reliable online templates and avoid poor-quality templates that lead to inferior tracking performance. In addition, the temporal information captured by the online template and the appearance information contained in the initial template reinforce each other to generate discriminative spatio-temporal features for target localization.

Specifically, the two branches of TATrack refer to the initial template and the online template respectively for multi-modal feature extraction and relation modeling via ViT \cite{dosovitskiy2010image}. Inside each branch, modality prompts are used to adjust the inputs of the transformer encoder and integrate feature extraction and cross-modal interaction. Furthermore, we design a spatio-temporal interaction (STI) mechanism based on the self-attention mechanism \cite{waswani2017attention}. STI enables cross-modal interaction to span longer time scales, instead of being limited to a pair of images. Spatio-temporal information can serve as a powerful guide for cross-modal interaction, aggregating object-oriented modality fusion features from search region for adaptive and precise information enhancement and complement. Finally, we fuse the feature maps of two branches and obtain the tracking results through a localization head.
The main contributions are summarized as follows:

\begin{itemize}
    \item We propose a temporal adaptive RGBT tracking framework, named as TATrack, which integrates feature extraction and cross-modal interaction with modality prompts and comprehensively exploit spatio-temporal and multi-modal information for RGBT tracking.
    \item We design a spatio-temporal interaction mechanism (STI). STI enables cross-modal interaction to span longer time scales and spatio-temporal information guides cross-modal interaction to generate more discriminative modality fusion features. 
    \item The proposed TATrack achieves state-of-the-art performance on three popular RGB-T tracking benchmarks, including RGBT210, RGBT234, LasHeR.
\end{itemize}

\begin{figure*}[ht]
    \centering
    \includegraphics[width=0.9\linewidth]{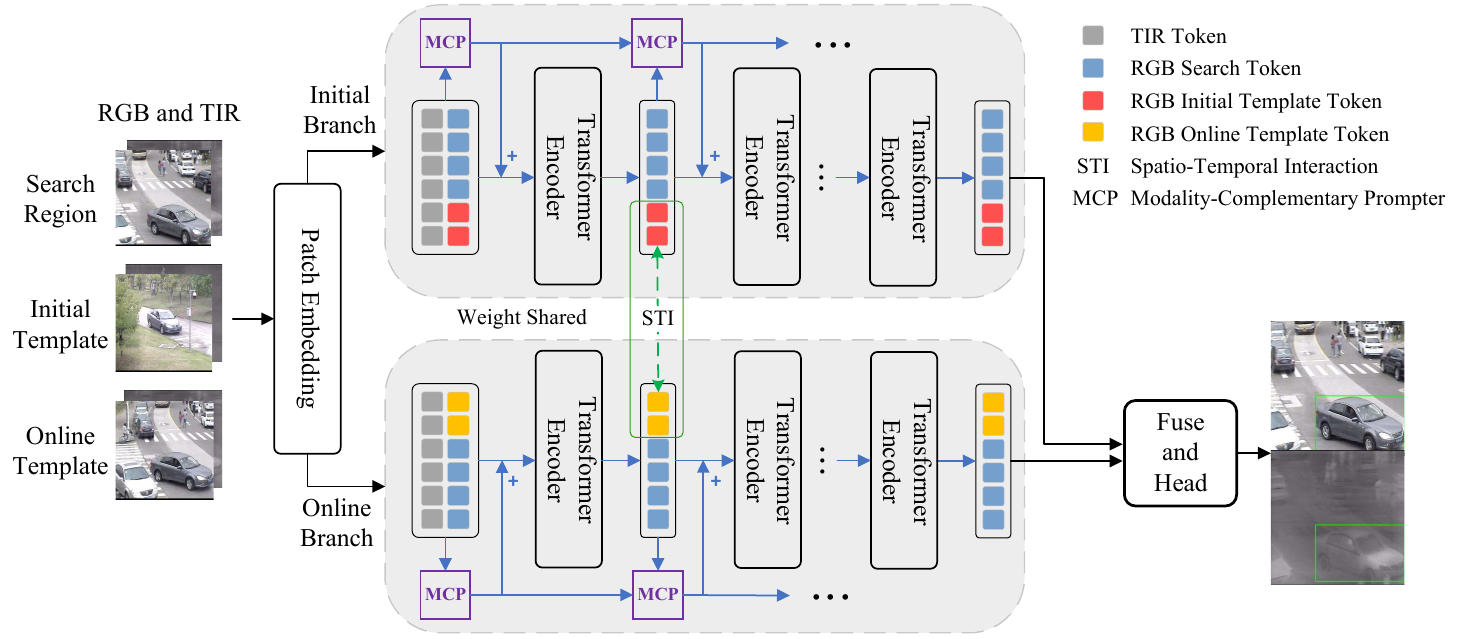}
    \caption{The overall framework of TATrack. The triplet of the search area, the initial template, and the online template is first embedded into tokens by the patch embed. The initial branch and the online branch respectively refer to the initial template and the online template for feature extraction and cross-modal interaction. The prompter generates modality prompts and adjusts inputs for the transformer encoder. STI enables the cross-frame propagation of spatio-temporal and multi-modal information.}
    \label{fig:2}
    \end{figure*}

\section{Related Works}

\subsection{Temporal Information Exploitation}

Tracking is a dynamic process and the state of objects is constantly changing over time. How to exploit temporal information to improve the robustness of trackers has been widely studied on VOT. For example, UpdateNet \cite{zhang2019learning} is proposed to estimate the optimal template for Siamese trackers \cite{bertinetto2016fully,zhu2018distractor}. LTMU \cite{dai2020high} learns a meta-updater to determine whether the tracker should be updated in the current frame. Stark \cite{yan2021learning} concatenates the initial template, the online template, and the search region to capture long-term dependencies in both spatial and temporal dimensions. Despite the strong similarity to VOT, RGBT tracking needs to consider the fusion of complementary information between two modalities, which limits the exploitation of temporal information. Prevailing RGBT trackers usually exploit temporal information through online training, and two classical works are correlation filter-based trackers \cite{wang2018learning,zhai2019fast} and MDNet-based trackers \cite{zhang2018learning,mei2023differential,xiao2022attribute}. The former mostly does not require offline training but samples and trains online. The latter requires offline training and update domain-specific layers during tracking. Although these methods are effective, they neglect the correlation of temporal information and spatial information. They are always accompanied by incorrect updates due to the lack of judgment on the reliability of the tracking results. In contrast, we capture temporal information by a reliable online updated template and learn spatio-temporal and modality information as a whole.

\subsection{Visual Prompt Learning}

In the field of NLP, ``pre-train, prompt'' has replaced ``pre-train, fine-tune'' as the dominant paradigm, which adapts the foundation model to different tasks by adding a textual prompt to the model inputs. When the pre-trained base model is trained on the downstream data, the fine-tuning will update all the parameters of the base model and the prompt-tuning will learn the task-oriented prompts. Prompt learning can achieve comparable performance to full fine-tuning even in the few-shot or zero-shot settings \cite{lester2021power,brown2020language,schick2020exploiting,schick2020s}, and significantly reduce memory usage and per-task storage. Some researchers transfer the new paradigm to computer vision and achieve promising performance. For example, CPT \cite{yao2021cpt} reformulates visual grounding into a fill-in-the-blank problem with color-based co-referential markers in image and text. \cite{bahng2022exploring} creates prompts with pixels demonstrating that prompts in pixel space indeed works successfully. VPT \cite{jia2022visual} applies visual prompts to vision backbones on 24 classification tasks. Convpass \cite{jie2022convolutional} introduces hard-coded inductive bias of convolutional layers to ViT in visual tasks. Besides, some works have demonstrated the potential of prompt learning in RGBT tracking. ProTrack \cite{yang2022prompting} first introduces prompts into the tracking field and proposes multi-modal prompts without the tuning process. ViPT \cite{zhu2023visual} learns the modal-relevant prompts to adapt the frozen pre-trained foundation model to various downstream multi-modal tracking tasks. Prompts can bridge RGB and TIR modalities in a simple and effective way. In this paper, we exploit modality prompts and attention module to enable cross-modal interaction to span longer time scales.

\section{Method}

In this section, we detail the proposed TATrack. First, we introduce a RGBT tracking baseline, which enables cross-modal interaction between visible and infrared images via modality prompts. The baseline disregards the state changes of objects and only refers to the initial template for matching-based tracking. Then, we extend the baseline to TATrack, comprehensively exploiting spatio-temporal information and multi-modal information for target localization. The overall architecture of TATrack is shown in Figure \ref{fig:2}.

\subsection{Tracking Baseline with Modality Prompt}

Inspired by the great success of ViT in vision tasks, the baseline is built on ViT. It mainly consists of three components: a ViT backbone, a modality-complementary prompter (MCP), and a bounding box prediction head.

The inputs of baseline are RGB and TIR search regions $I^r_x$, $I^t_x\in R^{3\times W_x\times H_x}$, RGB and TIR initial templates $I^r_{z}, I^t_{z}\in R^{3\times W_z\times H_z}$. They are first embedded into patches and flattened to 1D tokens $H_{rx}, H_{tx}\in R^{N_x\times C}$ and $H_{rz}, H_{tz}\in R^{N_z\times C}$, where $N_x=H_xW_x/P^2$, $N_z=H_zW_z/P^2$ ($P\times P$ is the resolution of each patch) and $C$ is the token dimension. We add learnable 1D position embeddings to the patch embeddings of the template and search region. Each pair of aligned images shares the same position embeddings. Then the token sequences are concatenated along the spatial dimension to $H^0_r=[H_{rz}, H_{rx}]$ and $H^0_t=[H_{tz}, H_{tx}]$. 

\textbf{Backbone.} The backbone includes $L$ standard visual transformer encoders for feature extraction and relation modeling. Each transformer encoder consists of Multi-head Self-Attention (MSA), LayerNorm (LN), Feed-Forward Network (FFN), and residual connection. The attention function is defined in equation \eqref{eq1}.

\begin{equation}
\begin{split}
A&{\rm =Softmax}(\frac{QK^T}{\sqrt C})V, \\
&={\rm Softmax}(\frac{H_rW_q(H_rW_k)^T}{\sqrt{C}})(H_rW_v),
\end{split}
\label{eq1} 
\end{equation}

\noindent where $\bf{Q, K, V}$ are query, key, and value matrices respectively. $W_q, W_k, W_v$ denote parameters of linear projections.

We denote $H^{l-1}_r$ as inputs to the $l$-th encoder layer $E^l$. The forward propagation process in the backbone is formulated as:
\begin{equation}
    {H^l_r=E^l(H^{l-1}_r),\quad \quad} l=1,2,...L
    \label{eq2}
\end{equation}

\textbf{Modality-Complementary Prompter.} ViPT \cite{zhu2023visual} proposes a modality-complementary prompter to generate valid visual prompts for the task-oriented multi-modal tracking. We generate infrared modality prompts through MCP. Modality prompts adjust the inputs of the transformer encoder so that the base model pre-trained on the large-scale RGB datasets is adapted to the downstream RGBT tracking task. The transformer encoder $E^l$ is equipped with a MCP module $P^l$. We denote $\mathbfcal{P}^{l}$ as the output to $P^l$. Modality prompts are generated as follows:
\begin{equation}
    {\mathbfcal{P}^l=P^l(\mathbfcal{P}^{l-1}, H^{l-1}_r),\quad } l=1,2,...L
    \label{eq3}
\end{equation}
\noindent $\mathbfcal{P}^0=H^0_t.$ MCP performs similar operations on the template sequence and the search sequence. Here we take the search sequence as an example to illustrate how the modality prompts are generated. First, MCP reshapes the two sequences to 2D features map and reduces their dimensions from $C$ to 8 by two $\rm 1\times 1$ convolutional layers. Then, the features of $H^{l-1}_r$ perform the spatial fovea operation and are added to features of $\mathbfcal{P}^{l-1}$. Next, the third $\rm 1\times 1$ convolutional layer is used to restore dimension for the mixed features. Then, MCP obtains modality prompts by flattening the mixed features. We refer readers to ViPT \cite{zhu2023visual} for more details about the modality-complementary prompter. Finally, modality prompts are added to the original inputs of the transformer encoder:
\begin{equation}
    {H^{l-1}_r=H^{l-1}_r+\mathbfcal{P}^l, \quad \quad} l=1,2,...L
    \label{eq4}
\end{equation}

Different from common RGBT trackers that implement cross-modal interaction through complex networks, modality prompts excavate inter-modal correlations and enable information complementation in a simple and effective way.

\textbf{Head.} A center head is placed to regress the bounding box. The sequence of search region is first reshaped into a 2D feature map and then fed into a fully convolutional neural network (FCN). The FCN has three branches, which are stacked by different numbers of Conv-BN-ReLU layers. The outputs of FCN contain the target classification score map, the local offset, and the normalized bounding box size. The bounding box is calculated by them.

\begin{figure}[t]
    \centering
    \includegraphics[width=0.9\linewidth]{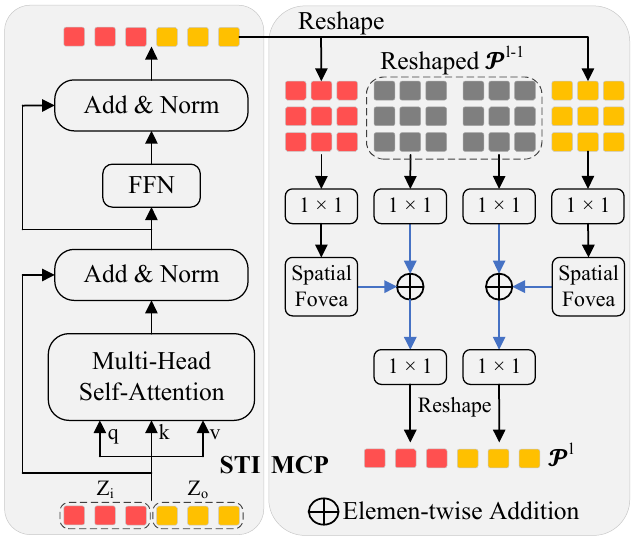}
    \caption{The processing in STI and MCP. TATrack gets a robust and precise representation of the target object by combining spatio-temporal information with multi-modal information.}
    \label{fig:3}
    \end{figure}

\subsection{Temporal Adaptive RGBT Tracking}

The baseline tracks only with reference to the initial template. However, during the tracking process, the state of a target object is constantly changing due to various factors such as target movement and environmental conditions. Capturing the latest state of the target object in time can significantly improve tracking performance. In this section, we describe in detail how to extend the baseline to TATrack.

\textbf{Overall Architecture.} TATrack captures temporal information by an online template that is dynamically updated during the tracking process, enabling TATrack to adapt to the state changes of the target object. Concretely, we copy the baseline (except the head) into two parallel branches, called the initial branch and the online branch. Following the process described earlier, the initial branch refers to the initial template for multi-modal feature extraction and cross-modal interaction, while the online branch refers to the online template. Between the two branches, STI achieves cross-frame propagation of spatio-temporal information. The search region sequences output from two branches are concatenated along the channel dimension. Then, we use a Conv-BN-ReLU layer to halve the channel and feed it into the head. 

\textbf{Spatio-temporal Interaction.} The STI aims to bridge the initial branch and the online branch to enable the propagation of spatio-temporal information across frames. Spatio-temporal information can serve as a powerful guide for multi-modal feature extraction and cross-modal interaction for adaptive and precise information enhancement and complementation. The STI performs information interaction between the template features of the two branches. We denote $Z_i$ and $Z_o$ as the template features output by one of transformer encoders in the initial branch and the online branch. The processing of STI and MCP is shown in Figure \ref{fig:3}.

The attention mechanism \cite{waswani2017attention} has been widely used in computer vision due to its powerful global modeling ability. Self-attention on concatenated sequences can realize information aggregation and interaction at the same time. That is, a bi-directional information flow between two sequences can be established by a single self-attention operation. We first concatenate the initial branch template features $Z_i$ and the online branch template features $Z_o$ along the spatial dimension, and then apply self-attention on the concatenated sequences for cross-frame information propagation:

\begin{equation}
\begin{gathered}
Z={\rm Concat}(Z_i, Z_o), \\
F={\rm Softmax}(\frac{Z\dot {W_q}(Z\dot {W_k})^T}{\sqrt{C}})(Z\dot {W_v}),
\end{gathered}
\end{equation}

\noindent where $\dot {W_q}, \dot {W_k}, \dot {W_v}$ denote parameters of the query, key, value projection layers. Then the spatio-temporal context $F$ and the concatenated sequences $Z$ are processed by LN, FFN, and residual connection as follows:

\begin{equation}
\begin{gathered}
\Tilde{F}{\rm =LN}(F + Z),\\
Z={\rm LN}(\Tilde{F}+{\rm FFN}(\Tilde{F})),
\end{gathered}
\end{equation}

STI establishes a bi-directional information flow that propagates spatial information in the initial template and temporal information in the online template to each other. The bi-directional information flow realizes the spatio-temporal interaction between the initial branch and the online branch. We insert the STI into the backbone before the modality-complementary prompter. These two templates aggregate spatio-temporal and multi-modal information respectively in STI and MCP for feature enhancement and complement, so that the tracker can capture the latest state of the target object. Robust and precise object representation enables the tracker to better deal with various challenges during matching-based tracking.

\textbf{Online Template Update.} TATrack captures temporal information by the online template to deal with state changes of the target object. However, poor-quality templates can mislead the tracker. Therefore, it is necessary to design a reasonable update scheme to pick reliable online templates. The maximum target classification score output by the prediction head reflects the confidence of the tracking result, which we use it as the basis for online template update. This scheme utilizes the discriminative ability of the model to the greatest extent without additional training.

\textbf{Differences with other prompt-learning methods.} Our work is inspired by visual prompt learning. There have been similar works applying the new paradigm to RGBT tracking. ProTrack \cite{yang2022prompting} first introduces the concept of prompt into RGBT tracking, but it only generates modality prompts through dyeing functions without tuning. ViPT \cite{zhu2023visual} generates learnable prompts and tunes on the downstream data to explore the associations between different modalities. But ViPT is a spatial-only tracker that disregards the state changes of objects. In contrast, TATrack is committed to exploring a new way to combine spatio-temporal and multi-modal information in RGBT tracking. Experiments show that TATrack achieves significant performance improvements.


\textbf{Training.} Our tracker is trained in an end-to-end fashion with the combination of classification and regression losses. We adopt the weighted focal loss \cite{law2018cornernet} for classification and L1 loss and the generalized IoU loss \cite{rezatofighi2019generalized} for bounding box regression as in \cite{ye2022joint}. The loss function can be written as:

\begin{equation}
    L_{track}=L_{cls}+\lambda_{iou}L_{iou}+\lambda_{L_1}L_1
\end{equation}

\noindent where $\lambda_{iou}$ = 2 and $\lambda_{L_1}$ = 5 are the regularization parameters. It is worth noting that during training, the parameters of the backbone and the prediction head are initialized and fixed as the same as the base model \cite{ye2022joint}, and we only update the other parameters.

\section{Experiments}

\subsection{Implementation Details}

TATrack is implemented in Python using PyTorch. The models are trained on 2 NVIDIA RTX 3090 GPUs and the inference speed is tested on a single NVIDIA RTX3090 GPU.

\textbf{Training.} We choose the LasHeR \cite{li2021lasher} dataset for fine-tuing our TATrack. Each GPU holds 32 image pairs, resulting in a global batch size of 64. The model fine-tuning takes 25 epochs, and each epoch contains 6 × $\rm{10^{4}}$ sample pairs. We train our model by AdamW optimizer \cite{loshchilov2017decoupled} with the weight decay $\rm{10^{-4}}$. The initial learning rate is set to 1 × $\rm{10^{-4}}$ and decreased by the factor of 10 after 10 epochs. The search regions and templates are resized to 128 × 128 and 256 × 256, respectively. TATrack is built on ViT-B, and STI is inserted in the 4-th, 7-th, and 10-th layers of the backbone. The fixed parameters are initialized with the base model \cite{ye2022joint} and the trainable prompt learning parameters are initialized with the Xavier uniform initialization scheme \cite{glorot2010understanding}.

\textbf{Inference.} We use the initial template, the online template, and the search region as inputs to TATrack. The online template is updated when the update interval of 50 is reached by default. The template with the maximum target classification score in the interval is selected to substitute the previous one.

\begin{table*}[!ht]
    \centering
    \begin{tabular}{c|cccccccccc|c}
    \hline
        ~ & HMFT & mfDiMP & CAT & MANet & MANet++ & MaCNet & APFNet & ProTrack & ViPT & TBSI & TATrack \\ \hline
        PR & 43.6 & 44.7 & 45.0 & 45.5 & 46.7 & 48.2 & 50.0 & 53.8 & 65.1 & 69.2 & \textbf{70.2} \\
        NPR & 38.1 & 39.5 & 39.5 & 38.3 & 40.4 & 42.0 & 43.9 & 49.8 & 61.7 & 65.7 & \textbf{66.7} \\
        SR & 31.3 & 34.3 & 31.4 & 32.6 & 31.4 & 35.0 & 36.2 & 42.0 & 52.5 & 55.6 & \textbf{56.1} \\ \hline
    \end{tabular}
    \caption{Evaluation results on LasHeR dataset.}
    \label{table:1}
\end{table*}

\begin{table*}[!ht]
    \centering
    \begin{tabular}{c|cccccccc|c}
    \hline
        ~ & MANet & MANet++ & APFNet & MaCNet & HMFT & mfDiMP & ProTrack & ViPT & TATrack \\ \hline
        NO & 67.2/46.3 & 63.6/40.7 & 66.7/46.7 & 74.0/51.7 & 77.8/55.5 & 76.5/57.5 & 75.4/58.0 & 84.0/68.4 & \textbf{88.4/71.3} \\ 
        PO & 42.4/30.7 & 44.0/30.1 & 47.3/34.5 & 44.6/32.8 & 38.4/27.7 & 39.7/30.8 & 50.5/39.6 & 62.4/50.3 & \textbf{67.6/53.9} \\ 
        TO & 35.0/26.0 & 35.4/25.4 & 41.7/31.4 & 38.6/29.2 & 30.8/22.2 & 32.2/25.0 & 43.9/34.2 & 57.6/46.1 & \textbf{62.2/49.3} \\ 
        HO & 24.1/23.6 & 24.5/24.4 & 27.1/27.7 & 28.1/29.1 & 19.6/21.5 & 19.8/23.8 & 40.2/38.6 & 43.7/43.8 & \textbf{48.7/45.0} \\ 
        MB & 38.9/27.9 & 39.7/26.6 & 45.9/32.8 & 40.4/29.8 & 37.5/26.2 & 37.6/28.7 & 52.4/39.5 & 57.3/45.9 & \textbf{62.8/49.9} \\ 
        LI & 35.6/26.9 & 35.8/24.0 & 41.8/30.8 & 36.0/26.7 & 33.0/24.5 & 29.6/23.8 & 42.4/33.4 & 49.8/41.2 & \textbf{54.3/44.0} \\ 
        HI & 47.3/34.4 & 53.3/34.7 & 60.4/41.2 & 52.0/37.4 & 48.4/34.7 & 46.7/35.1 & 59.5/44.4 & 67.9/54.2 & \textbf{75.6/59.8} \\ 
        AIV & 14.5/14.8 & 18.8/15.8 & 32.1/26.2 & 17.3/15.6 & 16.4/16.5 & 16.6/16.4 & 30.4/26.7 & 37.5/35.0 & \textbf{40.9/37.3} \\ 
        LR & 45.8/28.5 & 47.4/26.8 & 46.1/29.4 & 43.9/28.0 & 38.7/23.6 & 40.2/25.6 & 46.2/32.1 & 56.4/41.6 & \textbf{63.5/46.6} \\ 
        DEF & 37.4/32.1 & 39.4/30.8 & 45.8/36.8 & 41.4/34.0 & 34.9/27.5 & 40.3/34.2 & 51.9/42.8 & 67.4/55.7 & \textbf{75.6/61.4} \\ 
        BC & 38.3/30.2 & 43.6/31.4 & 44.9/33.7 & 42.2/31.9 & 35.4/25.9 & 34.9/27.0 & 49.8/38.8 & 64.9/51.8 & \textbf{67.4/53.5} \\ 
        SA & 38.0/27.9 & 41.1/27.9 & 42.8/31.7 & 40.8/30.4 & 36.1/26.5 & 37.2/29.5 & 45.1/36.3 & 57.3/46.5 & \textbf{62.6/50.3} \\ 
        CM & 42.8/31.2 & 42.2/29.4 & 47.7/35.1 & 46.7/33.9 & 42.5/29.4 & 40.8/30.6 & 54.1/41.6 & 62.1/50.0 & \textbf{67.6/53.6} \\ 
        TC & 38.6/27.3 & 40.1/26.8 & 43.1/31.6 & 39.8/28.7 & 36.3/25.5 & 38.0/28.8 & 45.8/35.8 & 57.3/46.0 & \textbf{61.9/49.4} \\ 
        FL & 30.2/19.4 & 37.8/21.6 & 37.6/27.9 & 34.6/22.2 & 31.3/21.7 & 32.3/25.7 & 52.0/38.6 & 59.1/46.5 & \textbf{70.6/55.2} \\ 
        OV & 32.1/34.9 & 28.0/22.0 & 36.4/34.2 & 34.8/36.7 & 41.9/36.0 & 40.6/34.9 & 54.8/45.8 & \textbf{76.2/65.0} & 73.1/63.2 \\ 
        FM & 41.0/30.6 & 41.1/28.9 & 45.1/33.9 & 43.7/33.0 & 39.6/29.3 & 41.3/32.4 & 52.0/41.4 & 63.1/51.4 & \textbf{68.8/55.3} \\ 
        SV & 46.0/32.9 & 46.4/31.1 & 49.8/36.0 & 48.0/34.8 & 43.8/31.7 & 45.2/34.9 & 54.5/42.5 & 65.0/52.5 & \textbf{70.1/56.1} \\ 
        ARC & 35.6/27.0 & 35.5/25.7 & 40.5/31.0 & 36.0/28.5 & 34.7/27.5 & 37.8/30.9 & 47.5/39.1 & 59.3/49.5 & \textbf{63.8/52.3} \\ \hline
        ALL & 45.5/32.6 & 46.7/31.4 & 50.0/36.2 & 48.2/35.0 & 43.6/31.3 & 44.7/34.3 & 53.8/42.0 & 65.1/52.5 & \textbf{70.2/56.1} \\ \hline
    \end{tabular}
    \caption{Attribute-based Precision/Success scores on LasHeR dataset.}
    \label{table:2}
\end{table*}

\subsection{Comparison with State-of-the-art Methods}

In this section, we compare the proposed TATrack with state-of-the-art RGBT trackers on three benchmarks including LasHeR \cite{li2021lasher}, RGBT234 \cite{li2019rgb}, and RGBT210 \cite{li2017weighted}.

\textbf{LasHeR.} LasHeR is a large-scale high-diversity benchmark for short-term RGBT tracking. LasHeR consists of 1224 visible and thermal infrared video pairs with more than 730K frame pairs in total. The test set contains 245 challenging video sequences. We compare our tracker with 10 other advanced trackers in terms of precision rate, normalized precision rate, and success rate. The comparison RGBT trackers include HMFT \cite{zhang2022visible}, MANet \cite{long2019multi}, mfDiMP \cite{zhang2019multi}, CAT \cite{li2020challenge}, MANet++ \cite{lu2021rgbt}, MaCNet \cite{zhang2020object}, APFNet \cite{xiao2022attribute}, ProTrack \cite{yang2022prompting}, ViPT \cite{zhu2023visual}, and TBSI \cite{hui2023bridging}. The results are reported in Table \ref{table:1}. It can be seen that our method outperforms previous SOTA methods by a large margin. TATrack exceeds ViPT by 5.1\%, 5.0\%, and 3.6\% in precision, normalized precision, and success, respectively, which proves that capturing the latest state of the target in time can significantly improve the tracking performance. Compared with the RGBT trackers trained online, the excellent performance of TATrack proves that capturing temporal information by an online updated template is a better way to exploit temporal information.

\textbf{RGBT234.} RGBT234 is a large-scale video benchmark dataset for RGBT tracking, which contains 234 video sequences totaling 234K frames. As shown in Table \ref{table:3}, TATrack outperforms the second place by 0.1\% and 0.7\% in precision and success, respectively. FANet \cite{zhu2020quality}, DAFNet \cite{gao2019deep}, JMMAC \cite{zhang2021jointly}, and CMPP \cite{wang2020cross} also participate in the comparison.

\textbf{RGBT210.} RGBT210 is a subset of RGBT234, which contains 210 video sequences totaling 210K frames. As shown in Table \ref{table:4}, TATrack outperforms CAT by 6.1\% and 8.5\% in precision and success, respectively. TATrack achieves comparable precision to TBSI, but falls behind in success  by 0.7\%. TFNet \cite{zhu2021rgbt} participates in the comparison.

\textbf{Attribute-Based Performance.} In order to evaluate the performance of TATrack in different scenarios, we also test it on sequences of different attributes in the LasHeR dataset. The attributes include no occlusion (NO), partial occlusion (PO), total occlusion (TO), hyaline occlusion (HO), motion blur (MB), low illumination (LI), high illumination (HI), abrupt illumination variation (AIV), low resolution (LR), deformation (DEF), background clutter (BC), similar appearance (SA), camera moving (CM), thermal crossover (TC), frame lost (FL), out-of-view (OV), fast motion (FM), scale variation (SV), and aspect ratio change(ARC). The results are shown in Table \ref{table:2}. TATrack has the best performance in almost all challenge attributes. The improvement is especially obvious in the scenarios of motion blur, high illumination, low resolution, deformation, camera moving, fast motion, and scale variation, which shows that TATrack can make good use of temporal information to deal with the state changes of the target while achieving modality complementary.

\begin{table}[!t]
    \centering
    \begin{tabular}{c|c|c}
    \hline
        Methods & Precision & Success \\ \hline
        mfDiMP & 64.6 & 42.8 \\ 
        DAFNet & 79.6 & 54.4 \\ 
        FANet & 78.7 & 55.3 \\ 
        MaCNet & 79.0 & 55.4 \\ 
        CAT & 80.4 & 56.1 \\ 
        JMMAC & 79.0 & 57.3 \\ 
        CMPP & 82.3 & 57.5 \\ 
        APFNet & 82.7 & 57.9 \\ 
        ProTrack & 79.5 & 59.9 \\ 
        ViPT & 83.5 & 61.7 \\
        TBSI & 87.1 & 63.7 \\ \hline
        TATrack & \textbf{87.2} & \textbf{64.4} \\ \hline
    \end{tabular}
    \caption{Evaluation results on RGBT234 dataset.}
    \label{table:3}
\end{table}

\begin{table}[!t]
    \centering
    \begin{tabular}{c|c|c}
    \hline
        Methods & Precision & Success \\ \hline
        TFNet & 77.7 & 52.9 \\ 
        mfDiMP & 78.6 & 55.5 \\
        CAT & 79.2 & 53.3 \\ 
        TBSI & \textbf{85.3} & \textbf{62.5} \\ \hline
        TATrack & \textbf{85.3} & 61.8 \\ \hline
    \end{tabular}
    \caption{Evaluation results on RGBT210 dataset.}
    \label{table:4}
\end{table}

\begin{table}[!t]
    \centering
    \begin{tabular}{c|ccc|c|c}
    \hline
        Model & MCP & STI & OTS & Precision & Success \\ \hline
        \ding{172} & ~ & ~ & ~ & 51.5 & 41.2 \\ 
        \ding{173} & \checkmark & ~ & ~ & 65.1 & 52.5 \\ 
        \ding{174} & \checkmark & \checkmark & ~ & 66.2 & 52.9 \\ 
        \ding{175} & \checkmark & ~ & \checkmark & 68.5 & 54.8 \\ \hline
        TATrack & \checkmark & \checkmark & \checkmark & 70.2 & 56.1 \\ \hline
    \end{tabular}
    \caption{Component analysis on LasHeR dataset. OTS: Online templates Select.}
    \label{table:5}
\end{table}

\subsection{Ablation Study}
To verify the effectiveness of the main components, we perform a detailed ablation study on the LasHeR dataset.

\textbf{Component Analysis.} As shown in Table \ref{table:5}, we compare four different models. \ding{172} denotes a single-branch RGB tracker consisting of a backbone and a localization head. \ding{173} denotes the baseline that adapts the RGB tracker to RGBT tracking by modality prompts. Compared with \ding{172}, the significant boost of \ding{173} proves the effectiveness of modality prompts. \ding{174} denotes that online template update is performed each frame while tracking. The performance degradation of \ding{174} compared to TATrack proves that TATrack selects reliable online templates. \ding{175} denotes the model with STI removed. Compared with TATrack, \ding{175} drops by 1.3\% and 1.7\% in success and precision, respectively, which illustrates the importance of spatio-temporal interaction.

\begin{table}[!ht]
    \centering
    \begin{tabular}{ccc|c|c|c}
    \hline
        \multicolumn{3}{c|}{Inserting Layers} & \multirow{2}{*}{Precision} & \multirow{2}{*}{Success} & \multirow{2}{*}{FPS} \\
        4 & 7 & 10 & ~ & ~ & ~ \\ \hline
        ~ & ~ & ~ & 68.5 & 54.8 & 28.3 \\ 
        \checkmark & ~ & ~ & 68.6 & 54.8 & 27.5 \\ 
        \checkmark & \checkmark & ~ & 69.5 & 55.6 & 26.8 \\ 
        \checkmark & \checkmark & \checkmark & 70.2 & 56.1 & 26.1 \\ \hline
    \end{tabular}
    \caption{Inserting layers of the proposed STI.}
    \label{table:6}
\end{table}

\begin{figure}[h]
    \centering
    \includegraphics[width=1.0\linewidth]{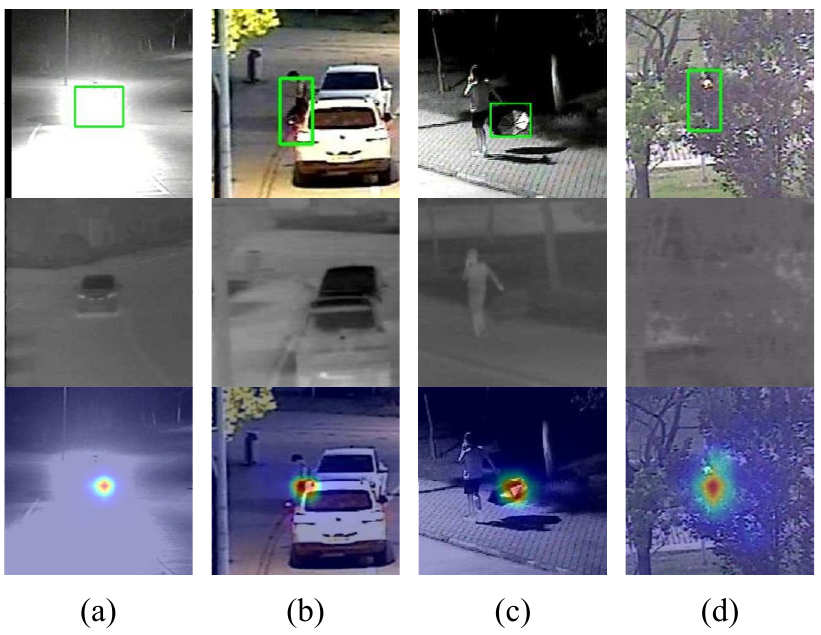}
    \caption{Visualization of response maps. The first row shows the RGB search region with a green bounding box. The second row shows the TIR search region. The third row shows the response map of the search region.}
    \label{fig:4}
    \end{figure}

\textbf{Inserting Layers of STI.} We experimentally investigate the effect of inserting layers of STI and summarize the results in Table \ref{table:6}. The performance improvement becomes more pronounced as the number of inserting layers in the backbone increases, which demonstrates the importance of spatio-temporal interaction. In addition, the insertion of STI makes the running speed decrease to some extent.

\textbf{Visualization.} To better illustrate the effectiveness of the proposed tracker, we visualize several representative response maps in Figure \ref{fig:4}. For example, the car in the first column is invisible in the RGB image due to the extreme illumination, but TATrack can make full use of the complementarity of RGB and TIR modalities to lock on the target stably. The umbrella in the third column is initially closed, and then the boy opens the umbrella causing a large deformation. Our method can capture the latest state of the target, thus TATrack is well focused on the target. In addition, our method can cope well with challenges such as occlusion, fast movement, scale change, and aspect ratio change.

\section{Conclusion}
In this paper, we propose the TATrack, a temporal adaptive RGBT tracking framework.  The core idea of TATrack is to explore a new way to make better use of temporal information in RGBT tracking. Different from online-trained trackers, TATrack captures temporal information by an online updated template and combines spatio-temporal information and multi-modal information for enhancement and complement. Extensive experiments on three RGBT tracking benchmarks show that our method achieves state-of-the-art performance. In the future, we will explore more template update options to determine whether the tracker should be updated at the moment.

\section*{Acknowledgments}
This work is supported by National Defense Basic Scientific Research Project JCKY2021208B023, Guangzhou Key Research and Development Program 202206030003, and Guangdong High-level Innovation Research Institution Project 2021B0909050008.

\bibliography{aaai24}

\end{document}